___

# Fast On-line signature recognition based on VQ with time modeling[1]

Juan Manuel Pascual-Gaspar (a), Marcos Faundez-Zanuy (b,*), Carlos Vivaracho (c)

(a) Valladolid, SPAIN
E-mail: pascualgaspar@gmail.com

(b) Escola Universitària Politècnica de Mataró, Barcelona, SPAIN
E-mail: faundez@eupmt.es

(c) Facultad de informática, Universidad de Valladolid, SPAIN
E-mail: cevp@infor.uva.es

**ABSTRACT**

This paper proposes a multi-section vector quantization approach for on-line signature recognition. We have used the MCYT database, which consists of 330 users and 25 skilled forgeries per person performed by 5 different impostors. This database is larger than those typically used in the literature. Nevertheless, we also provide results from the SVC database.
Our proposed system outperforms the winner of SVC with a reduced computational requirement, which is around 47 times lower than DTW. In addition, our system improves the database storage requirements due to vector compression, and is more privacy-friendly as it is not possible to recover the original signature using the codebooks. Experimental results with MCYT provide a 99.76% identification rate and 2.46% EER (skilled forgeries, individual threshold). Experimental results with SVC are 100% of identification rate and 0% (individual threshold) and 0.31% (general threshold) when using a two-section VQ approach.

## 1. INTRODUCTION

Handwritten signatures have a long tradition of use in commonly encountered verification tasks such as financial transactions and document authentication. They are easily used and well accepted by the general public, and signatures are straightforward to obtain with relatively cheap devices. These are important advantages of signature recognition over other biometrics. Yet, signature recognition also has some drawbacks: it is a difficult pattern recognition problem due to possible large variations between different signatures made by the same person. These variations may originate from instabilities, emotions, environmental changes, etc, and are person dependent. In addition, signatures can be forged more easily than other biometrics.

The signature recognition task can be split into two categories depending on the data acquisition method:

- Off-line (static), the signature is scanned from a document and the system recognizes the signature, analyzing its shape (Tylan, 2009; Frias-Martinez, 2006).

---

[1] This work has been supported by FEDER and MEC, TEC2009-14123-C04-04

_______________________________________________________________

- On-line (dynamic), the signature is acquired in real time by a digitizing tablet and the system analyzes shape and the dynamics of writing, using for example: position with respect to the x and y axes, pressure applied by the pen, etc.

Using dynamic data, further information can be extracted, such as acceleration, velocity, curvature radius, etc. (Plamondon, 1989). In this paper, we will focus on the online (dynamic) signature recognition task.

For a signature verification system, depending on the test conditions and environment, three types of forgeries can be established (Plamondon, 1989):

- Simple forgery, where the forger makes no attempt to simulate or trace a genuine signature.
- Substitution or random forgery, where the forger uses his/her own signature as a forgery.
- Freehand or skilled forgery, where the forger practices imitating as closely as possible the static and dynamic information of the signature to be forged.

From the point of view of security, the last one is the most damaging and, for this reason, some databases suitable for system development include some trained forgeries (Ortega-Garcia et al, 2003; Yeung et al, 2004).

The remaining sections of this paper will be devoted to the task of dynamic signature recognition. Section 2 looks at our proposal in greater detail, showing the similarities and differences with other related works. The experimental setup is shown in Section 3. The results can be seen in Section 4, as well as a comparison with other proposals. Finally, the conclusions and future work can be seen in Section 5. In the Appendixes show the VQ and DTW algorithms (appendix 1 and 2 respectively) and the computation burden comparison between our proposal and the state-of-the-art system (appendix 3).

## 2. SIGNATURE RECOGNITION BASED ON VQ

In this section we present several vector quantization algorithms for on-line signature recognition.

It has recently been found that the Dynamic Time Warping algorithm outperforms HMM for signature verification (Houmani et al., 2009). For this reason, we will use DTW as the baseline algorithm for performance comparison.

Appendix 2 describes the DTW algorithm. In our case, we will use five signatures per person, acquired during enrollment. DTW works out the distance between the test set of vectors and each of the five training sets.

Distance computation implies a warping by means of dynamic programming. We compute five distances, each one being the result of comparing the test sequence with each of the five training repetitions. These five distances are combined using different approaches, such as *min*{}, *mean*{}, and *median*{}, in order to obtain the final distance. A more detailed explanation of VQ and DTW algorithms can be found in (Faundez-Zanuy, 2007).

___

Template matching approaches are especially appropriate when a small number of samples are available for training a model. This is the case with on-line signature recognition. VQ (Faundez-Zanuy, 2007) does not model the temporal evolution of the signature because, when averaging the vectors, the time ordering is lost (the vectors are all mixed together, discarding the temporal instant of their generation). This is a drawback that can be solved in at least a couple of ways that we introduce next:

a) The inclusion of temporal information of feature vectors: one of the components of the feature vector is the time instant of acquisition, as will be shown in section 3.3, table 2.
b) Using a multi-section approach: This method is an improved version of the classical vector quantization approach proposed in (Faundez-Zanuy, 2007), which can also be interpreted as a variant of the split-VQ (Gersho and Gray, 1991). This proposal is described in the next section.

## 2.1 Proposed algorithm based on Multi-section Codebook approach

DTW offers one advantage over VQ: it takes into account the temporal evolution of the signature. However, a simple model called the multi-section codebook (Burton et al., 1983; Buck et al., 1985) was proposed in the mid-eighties in speech and speaker recognition. Although this approach was discarded due to the higher accuracies of HMM, we should take into account the fact that signature recognition differs from speech/ speaker recognition, as the length of the training set is rather short and it is hard in this situation to estimate an accurate statistical model. This observation is well known in the field of speaker recognition, where higher recognition rates using VQ, as compared with HMM, have been reported for short training/ testing sets.

The multi-section codebook approach consists of splitting the training samples into several sections. For example, figure 1 represents a three-section approach, where each signature is split into three equal length parts (initial, middle and final sections). In this case, three codebooks must be generated for each user, each codebook being adapted to one portion of the signature. Each branch works in a similar fashion to the VQ approach, and the final decision is taken by combining individual contributions of each section by simple averaging.

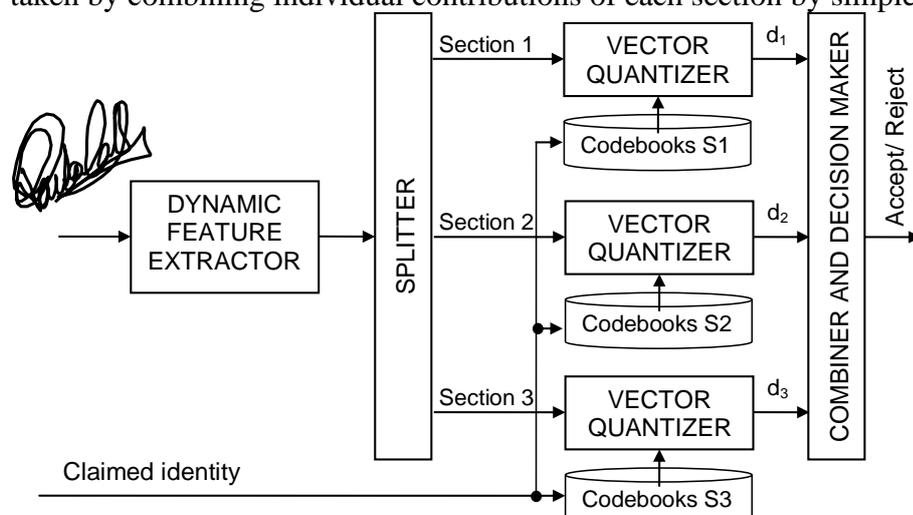

**Figure 1.** A multi-section codebook approach for signature verification based on 3 sections.

If the database contains *P* users and the splitter provides *S* sections we will have *S* codebooks for each person. Thus, we will have one codebook per person and section, named $CB_{p,s}$ for *p*=1,..,*P* and *s*=1,…,*S*.

This proposal is a generalization of the VQ approach, which can be seen as a multi-section approach with just a single section. The multi-section system will be operated as follows.

*2.1.1 User model computation*

For each person, we split the signature into *S* sections. We concatenate the feature vectors resulting from each of the five training signatures belonging to the same section. Thus, we obtain *S* training sequences per person. In order to obtain one codebook per person and section, we have applied a codebook initialization plus Lloyd iteration for codebook improvement. Codebook initialization is obtained by splitting the signature in as many segments of equal length as the final number of centroids. One centroid per portion is obtained as the average of the points belonging to that segment. Our experimental results confirm that this approach offers similar results to the classic LBG algorithm, but with less computation time.

Each person is thus modeled by a set of *S* codebooks. Figure 2 schematically represents the process of splitting and generating the training sequences for a given user *p*. It is assumed that five signatures of the same person are used for training the user model.

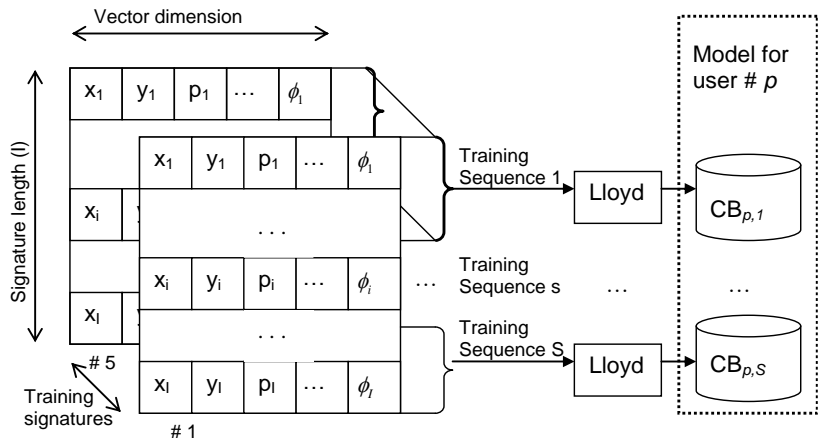

**Figure 2.** Schematic representation of the procedure followed to obtain the training sequences and the codebooks for a given user *p*.

This strategy is similar to a constrained vector quantization approach named split-VQ (also known as partitioned VQ) (Gersho et al., 1991). The simplest and most direct way to reduce the search and storage complexity in coding a high dimensional vector is simply to partition the vector into two or more sub-vectors. However, rather than splitting vectors, we split the training and testing sequences into sections, which have the same original vector dimension.

*2.1.2 User recognition*

_________________________________________________________________

The quantization distortion for a given signature and person $p$ must be obtained by computing the combination of the distortions obtained for each part. This can be achieved by a generalization of the procedure described in (Faundez-Zanuy 2007): $dist_p = combination(d_1,...,d_S)$. The most straightforward combination is the *mean{}* function. However, several combinations have been evaluated, as described in the next section.

The individual $d_s$ values, for $s=1,..,S$ and person $p$, are obtained by applying the equation

$$d_s = \sum_{i=1}^{I'} NNER(\vec{x}_i, CB_{p,s}) = \sum_{i=1}^{I'} \min_{\vec{y}_j \in CB_{p,s}} \{d(\vec{x}_i, \vec{y}_j)\}$$

where NNER is the Nearest Neighbor Encoding Rule (Gersho and Gray, 1991). It is interesting to point out that $I' \cong \dfrac{I}{S}$ as we split the whole signature into $S$ sections of equal length.

*2.1.3 Multi-section distance fusion*

The combination techniques evaluated in the combiner block of figure 1 are as follows:

- Minimum value (min): $dist_p = \min(d_1,...,d_S)$
- Maximum value (max): $dist_p = \max(d_1,...,d_S)$
- Sum (S): $dist_p = \sum_{i=1}^{S} d_i$. It is equivalent to the *mean{}* function, since the number of sections is a fixed value.
- Product (P): $dist_p = \log(\prod_{i=1}^{S} d_i) = \sum_{i=1}^{S} \log d_i$. To avoid small values, the *log* function was used.
- Sum using Extreme Values (SEV): $dist_p = \min(d_1,...,d_S) + \max(d_1,...,d_S)$. The combination of extreme values, which has shown a good performance in previous works (Vivaracho et al., 2003), was then tested here.
- Weighted sum, which can be with user independent weighting: $dist_p = \sum_{i=1}^{S} c_i \times d_i$ and user dependent weighting: $dist_p = \sum_{i=1}^{S} c_{p_i} \times d_i$. Both have been tested as follows:
  - Based on training samples statistics. Once the model for the person $p$ is trained, the distances with regard to each training sample section is calculated. Let us call $d_{p_i}^{t_j}$ the distance for section $i$ of the training sample $t_j$. From these values, each section mean, $\mu_{p_i}$, and standard deviation, $\sigma_{p_i}$ are calculated as follows:

$$\mu_{p_i} = \frac{\sum_{j=1}^{T} d_{p_i}^{t_j}}{T} \quad 1 \leq i \leq S$$

$$\sigma_{p_i} = \sqrt{\frac{\sum_{j=1}^{T}(d_{p_i}^{t_j} - \mu_{p_i})^2}{T}} \quad 1 \leq i \leq S$$

  where $T$ is the number of samples used to train the model. Based on these statistics, the following combinations were tested:
    - Weighted Sum based on Deviation (WSD). The standard deviation can be considered as a measure of dispersion, then, the smaller it is, the more stable the section score will be. Under this assumption, the user weighting coefficients must be inversely proportional to $\sigma_{p_i}$:

___________________________________________________________________

$$c_{p_i} = \frac{1/\sigma_{p_i}}{\sum_{k=1}^{S} 1/\sigma_{p_i}} \quad and \quad \sum_{i=1}^{S} c_{p_i} = 1$$

- Weighted Sum based on High Means (WSHM). Those sections with high scores (high means) are reinforced:

$$c_{p_i} = \frac{\mu_{p_i}}{\sum_{k=1}^{S} \mu_{p_k}} \quad and \quad \sum_{i=1}^{S} c_{p_i} = 1$$

- Weighted Sum based on Low Means (WSLM). Those sections with low scores (low means) are reinforced:

$$c_{p_i} = \frac{1/\mu_{p_i}}{\sum_{k=1}^{S} 1/\mu_{p_i}} \quad and \quad \sum_{i=1}^{S} c_{p_i} = 1$$

o Based on recognition errors. The idea is to reinforce those sections with better performance. Several implementations of this idea have been tested:
- Weighted Sum based on Random forgery Errors (WSRE). By means of a development set, each section *i* performance, measured by means of the Equal Error Rate ($EER_i$), is calculated using random forgeries. Those sections with better performance (lower errors) are reinforced:

$$c_i = \frac{1/EER_i}{\sum_{k=1}^{S} 1/EER_i} \quad and \quad \sum_{i=1}^{S} c_i = 1$$

- Weighted Sum based on Skilled forgery Errors (WSSE). The same as the previous one but using skilled forgeries to compute $EER_i$.
- Weighted Sum based on User dependent Errors (WSUE). Here, we cannot use skilled forgeries since, in a real application, we cannot imitate the user signature due to legal and social restrictions. To calculate the $EER_{p_i}$, the training sample distances, $d_{p_i}^{t_j}$, are the target scores and the distances between other user training samples and the user model are the impostors. The weighting coefficients are calculated in the same way as in WSRE, but here the coefficients are user dependent, $c_{p_i}$.
- Weighted Sum based on User dependent EER Thresholds (WSUT). The problem with the previous approach is that most of the $EER_{p_i}$ are 0, fixing the corresponding $c_{p_i}$ at 1. It was observed, by analyzing the system performance, that, in general, low errors were associated with low EER thresholds and vice versa. Then, the use of the threshold instead of the EER was tested.

## 3. EXPERIMENTAL SETUP

### 3.1 Databases

Experimental results, obtained with two publicly available databases, have been achieved.

### 3.1.1 MCYT Database

We used our previous database MCYT (Ortega-Garcia et al., 2003), acquired with a WACOM graphics tablet. The sampling frequency for signal acquisition is set to 100 Hz, yielding the following set of information for each sampling instant:

(i) position along the x-axis, $x_t$ : [0–12 700], corresponding to 0–127 mm;

___________________________________________________________________

(ii) position along the y-axis, $y_t$ : [0–9700], corresponding to 0–97 mm;
(iii) pressure $p_t$ applied by the pen: [0–1024];
(iv) azimuth angle $\gamma_t$ of the pen: [0–3600], corresponding to 0 –360;
(v) altitude angle $\varphi_t$ of the pen: [300–900], corresponding to 30 –90;

We have used extended feature vectors from these five measurements. We recruited 330 different users. Each target user produces 25 genuine signatures, and 25 skilled forgeries are also captured for each user. These skilled forgeries are produced by the 5 subsequent target users by observing the static images of the signature to be imitated, trying to copy them (at least 10 times), and then, producing valid forgeries in a relaxed fashion (i.e. each user acting as a forger is requested to sign naturally, without artefacts, such as breaks or slowdowns). In this way, highly skilled forgeries with shape-based natural dynamics are obtained. Following this procedure, user n (ordinal index) carries out a set of 5 samples of his/her genuine signature, and then 5 skilled forgeries of client n–1. Then, again, a new set of 5 samples of his/her genuine signature; and then 5 skilled forgeries of user n–2. This procedure is repeated by user n, making further samples of the genuine signature and imitating previous users n–3, n–4 and n–5. Summarizing, user n produces 25 samples of his/her own signature (in sets of 5 samples) and 25 skilled forgeries (5 forgeries of each user, n–1 to n–5). In a similar way, for user n, 25 skilled forgeries will be produced by users n+1 to n+5.

### 3.1.2 SVC Database

The SVC database (Yeung et al., 2004) is very similar to MCYT. The sub-database released for Task 2 of the First international signature verification competition also includes the same five features as MCYT acquired by a WACOM Intuos graphic tablet with a sampling rate of 100Hz. The complete SVC database had 100 sets (users) of signature data, but just one subset of 40 users was made available for research after the competition.

This database also contains skilled forgery samples produced by the contributors. There are 20 genuine signatures per user collected through two sessions, 10 signatures per session, with a minimum of one week between sessions. Additionally, there are 20 skilled forgeries produced by at least four other contributors. The skilled forger was provided with a software animation viewer of the signature to be forged. Thus, in this work, we have used a final set of 16,000 signatures (8,000 genuine signatures plus 8,000 skilled forgeries). This is about 10% of the MCYT database size.

It must be pointed out that the signatures in the SVC database are mostly in either English or Chinese, and no 'real' signatures were used. Instead, the contributors were advised to design a new signature and practice it before the acquisition sessions.

### 3.2 Conditions of the experiments

The databases have been split into the following sub corpora:

- Development Set (DS). This is used to optimize the parameters of the system. It consists of 80 people from the MCYT database.

___________________________________________________________________

- Test Set 1 (TS1). It consists of the whole MCYT database except for users of DS, i.e., 250 people.
- Test Set 2 (TS2). It consists of a subset of the SVC 2004 that was made available for research after the competition, i.e., 40 people.

Training and testing signatures have been chosen in the following way:

- **MCYT database**
    - We performed identification experiments, using the first 5 signatures per person for training and 5 different signatures per person for testing (signatures 20 to 24). This implies a total number of 80×5 tests for the DS and 250×5 tests for the TS.
    - We performed verification experiments, using the first 5 signatures per person for training and 20 different genuine signatures per person for testing (signatures 6 to 24). In addition, we used the 25 available skilled forgeries made by 5 other users and 5 genuine signatures from other signatories for impostor tests. This implies a total number of 80×20 genuine tests plus 80×25 impostor tests (skilled forgeries) and 80×79×5 impostor tests (random forgeries) for the DS and 250×20 genuine tests plus 250×25 impostor tests (skilled forgeries) and 250×249×5 impostor tests (random forgeries) for the TS1.
- **SVC database**
    - We performed identification experiments, using the first 5 signatures per person for training and 5 different signatures per person for testing (signatures 16 to 20). This implies a total number of 40×5 tests.
    - We performed verification experiments, using 5 signatures per person for training and 15 different genuine signatures per person for testing. In addition, we used the 20 skilled forgeries made by other users and 5 genuine signatures from other signatories for impostor tests. This implies a total number of 40×15 genuine tests plus 40×20 impostor tests (skilled) and 40×39×5 impostor tests (random).

However, further study needs to be done on whether this database can produce statistically significant results. In (Guyon et al., 1998), the minimum size of the test data set $N$, which guarantees statistical significance in a pattern recognition task, is derived. The goal in this work is to estimate $N$ so that it is guaranteed, with a risk α of being wrong, that the error rate $P$ does not exceed the estimation $\hat{P}$ from the test set by an amount larger than $\varepsilon(N,\alpha)$; that is,

$$\Pr\{P > \hat{P} + \varepsilon(N,\alpha)\} < \alpha$$

Letting $\varepsilon(N,\alpha) = \beta P$, and supposing recognition errors to be Bernoulli trials (i.i.d. errors), after some approximations, the following relationship can be derived:

$$N \approx \frac{-\ln \alpha}{\beta^2 P}$$

For typical values of α and β (α =0.05 and β =0.2), the following simplified criterion is obtained:

$$N \approx \frac{100}{P}$$

___

If the samples in the test data set are not independent (due to correlation factors that may include variations in recording conditions, in the type of sensors, etc.), then *N* must be further increased. The reader is referred to (Guyon et al., 1998) for a detailed analysis of this case, where some guidelines for computing the correlation factors are also given.

Table 1 shows the number of tests done in each condition and, with 95% confidence, the statistical significance in experiments with an empirical error rate, down to $\hat{P}$. Thus, the experiments of this section are statistically significant, because our errors are higher than those presented in table 1.

**Table 1.** Statistical significance in experiments, with 95% confidence.

|  | MCYT (250 users) | MCYT (80 users) | SVC Data Base (40 users) |
|---|---|---|---|
| Random forgeries | $\hat{P}$ =0.03% | $\hat{P}$ =0.3% | $\hat{P}$ =1.25% |
| Skilled forgeries | $\hat{P}$ =0.89% | $\hat{P}$ =2.78% | $\hat{P}$ =10% |

### 3.3 Feature selection

The selection of an appropriate combination of features is fundamental in signature recognition, so the initial set of features provided by the tablet was optimized to improve the output of our system. We compared different combinations of features in the domain of the position and velocity, which have been shown to be the most decisive for signature recognition (Pascual-Gaspar et al., 2009).

We calculated the center of mass of each signature and displaced this point to the origin of the coordinates.

Table 2 shows the results of the preliminary tests for feature selection. These tests were made with a codebook (single section) of 5 bits, and both substitution and skilled impostors. The symbols in the first column of the table have the following interpretation:

- x, y: geometric coordinates
- p: pressure
- γ, φ: angular features
- dx, dy, dp, dγ, dφ: first temporal derivatives of previous features
- t: timestamp

All features, including timestamp, were normalized through a statistical preprocessing based on the standard z-norm.

______________________________________________________________________

**Table 2.** Experiments for optimal feature set selection with MCYT. Size of the CB is 5 bits.

| Feature set | EER (%) | |
|---|---|---|
| | substitution | Skilled |
| $FS_1=[x,y,p,\gamma,\varphi]$ | 5.07% | 13.24% |
| $FS_2=[x,y,dx,dy]$ | 0.85% | 4.28% |
| $FS_3=[x,y,p,dx,dy,dp]$ | 0.39% | 4.02% |
| $FS_4=[x,y,p,dx,dy]$ | 0.51% | 3.56% |
| $FS_5=[x,y,dx,dy,dp]$ | 0.21% | 4.25% |
| $FS_6=[x,y,dx,dy,dp,t]$ | 0.19% | 2.55% |

As can be seen, with the same codebook configuration, results in error terms are strongly dependent on the features used. The feature set $FS_1$, the default provided by the tablet, is significantly the worst of all the sets. Because of its better performance with both types of impostor, the combination finally selected was $FS_6$.

### 3.4 Performance measure

In identification, the % of signatures correctly assigned (% of success) will be shown.

Verification systems can be evaluated using the false acceptance rate (FAR, those situations where an impostor is accepted) and the false rejection rate (FRR, those situations where a user is incorrectly rejected), also known in detection theory as false alarm and miss, respectively. A trade-off between both errors usually has to be established by adjusting a decision threshold. The performance can be plotted on an ROC (receiver operator characteristic) or a DET (detection error trade-off) plot.

However, for a better system performance comparison, the use of a single number measure is more useful and easier to understand. One of the most popular is the Equal Error Rate (EER), that is, the error of the system when the decision threshold is such that the FMR equals the FNMR. The EER will be the measure used in this paper.

The EER can be evaluated with a different threshold for each user (Individual threshold, I) or with the same for all (General threshold, G). The latter is the less favourable case, due to the variability of the scores from one user to another. The first can be considered as the lower limit of the second. EER has been evaluated with both thresholds.

### 4. Experimental results

### 4.1 Verification task

### 4.1.1 Development Set

Fig. 3 shows the evolution of the results (EER in %) with respect to the codebook size for the single section VQ. Codebook sizes of 1, 2, 4, 8, 16, 32, 64, 128, 256 and 512 were tested.

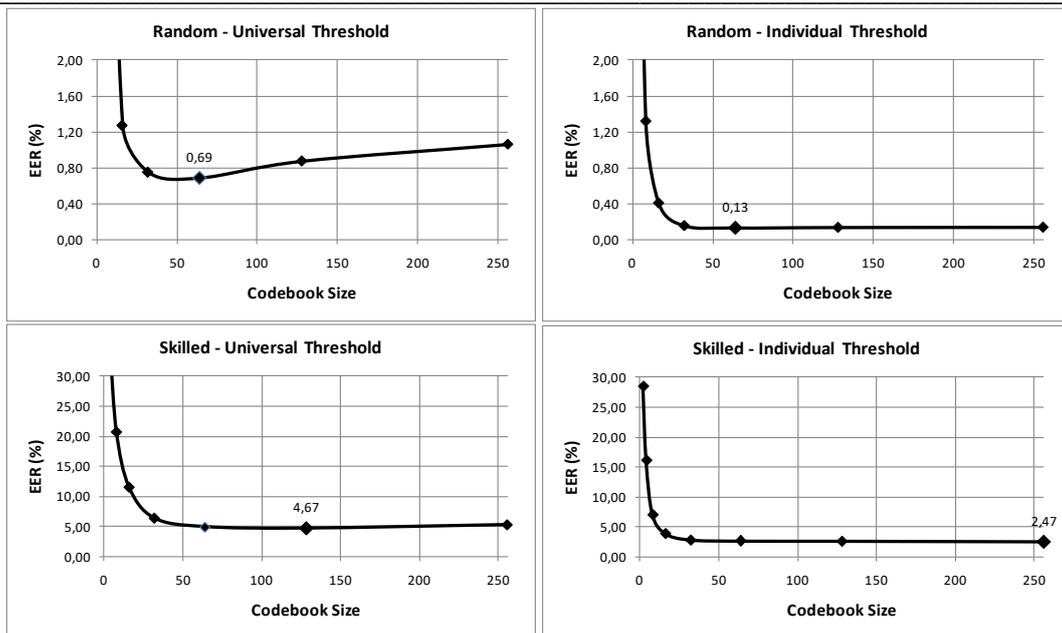

**Figure 3.** Results achieved with the VQ single section for the development set. The best results are emphasized and the corresponding EER (in %) is included.

Table 3 shows the results with the multi-section approach. Multi-section sizes from 2 to 8 were tested. Column *Sec* shows the number of sections, *FT* shows the Forgery (R: Random and S: Skilled) and Threshold (I: Individual and G: General) types. The rest of the columns show the fusion techniques tested (see Sec. 2.1.3). The row *CB* shows the optimal codebook size (that with the best performance) for each number of sections and combination technique. The best results are emphasized in bold.

The best results with multi-section are achieved with 2-3 sections, being close to those obtained with a single VQ. The greater the number of sections, the worse the results. With respect to the combination techniques, the weighted sum ones proposed work well, but do not significantly outperform the sum and product techniques.

___

**Table 3.** EER (in %) obtained in the multi-section experiments with the development set.

| Sec | FT | min | max | S | P | SEV | WSD | WSHM | WSLM | WSRE | WSSE | WSUE | WSUT |
|---|---|---|---|---|---|---|---|---|---|---|---|---|---|
| CB |  | 128 | 64 | 32 | 64 | 32 | 64 | 32 | 32 | 64 | 64 | 64 | 32 |
|  | RI | 0.8 | 0.6 | 0.2 | 0.2 | 0.2 | 0.2 | 0.3 | 0.2 | 0.2 | 0.2 | 0.2 | 0.2 |
| 2 | RG | 2.2 | 1.5 | **0.9** | 1.1 | **0.9** | 1.1 | 1.0 | **0.9** | 1.1 | 1.1 | 1.1 | **0.9** |
|  | SI | 4.6 | 4.2 | 3.1 | 2.9 | 3.1 | 3.0 | 3.2 | 3.2 | 2.9 | 3.1 | 3.1 | 3.2 |
|  | SG | 7.0 | 6.2 | **4.8** | **4.8** | **4.8** | 5.0 | **4.8** | 5.3 | **4.8** | 4.9 | **4.8** | 4.9 |
| CB |  | 256 | 64 | 32 | 32 | 32 | 32 | 32 | 32 | 32 | 32 | 32 | 32 |
|  | RI | 1.5 | 0.7 | 0.2 | 0.3 | 0.4 | 0.3 | 0.2 | 0.3 | **0.17** | 0.3 | 0.2 | 0.3 |
| 3 | RG | 3.7 | 2.0 | 1.1 | 1.1 | 1.2 | 1.1 | 1.0 | 1.2 | 1.0 | 1.1 | 1.1 | 1.2 |
|  | SI | 4.5 | 4.3 | 3.0 | 3.0 | 3.1 | 3.0 | 3.1 | 3.1 | **2.8** | 3.1 | 2.9 | 3.1 |
|  | SG | 7.7 | 8.1 | 5.0 | 4.9 | 5.4 | 5.2 | 5.2 | 5.4 | **4.8** | 5.2 | 5.5 | 5.3 |
| CB |  | 64 | 64 | 32 | 32 | 32 | 32 | 32 | 16 | 32 | 16 | 32 | 32 |
|  | RI | 2.1 | 1.1 | 0.3 | 0.4 | 0.7 | 0.6 | 0.3 | 0.5 | 0.3 | 0.4 | 0.4 | 0.4 |
| 4 | RG | 4.1 | 2.4 | 1.3 | 1.5 | 1.7 | 1.7 | 1.4 | 1.5 | 1.3 | 1.4 | 1.5 | 1.6 |
|  | SI | 5.5 | 4.9 | 3.3 | 3.3 | 3.5 | 3.8 | 3.4 | 3.8 | 3.2 | 3.6 | 3.3 | 3.7 |
|  | SG | 7.8 | 8.7 | 5.2 | 5.4 | 6.3 | 6.0 | 5.6 | 5.6 | 5.2 | 5.4 | 5.4 | 5.6 |
| CB |  | 64 | 32 | 16 | 32 | 16 | 16 | 32 | 16 | 16 | 32 | 64 | 16 |
|  | RI | 2.3 | 1.5 | 0.3 | 0.4 | 0.8 | 0.5 | 0.4 | 0.5 | 0.3 | 0.4 | 0.3 | 0.5 |
| 5 | RG | 4.8 | 2.2 | 1.5 | 1.5 | 1.7 | 2.0 | 1.6 | 1.7 | 1.4 | 1.6 | 1.8 | 1.7 |
|  | SI | 5.7 | 5.7 | 3.3 | 3.3 | 4.2 | 4.1 | 3.2 | 3.7 | 3.1 | 3.4 | 3.2 | 4.0 |
|  | SG | 9.2 | 9.1 | 5.4 | 5.6 | 7.0 | 7.2 | 5.7 | 6.6 | 5.1 | 5.4 | 6.1 | 6.7 |
| CB |  | 32 | 16 | 32 | 32 | 16 | 16 | 16 | 16 | 64 | 16 | 64 | 16 |
|  | RI | 2.5 | 1.5 | 0.5 | 0.5 | 1.0 | 0.6 | 0.4 | 0.6 | 0.3 | 0.6 | 0.4 | 0.6 |
| 6 | RG | 4.4 | 1.7 | 1.8 | 1.8 | 2.0 | 1.7 | 1.6 | 1.9 | 1.6 | 1.8 | 1.9 | 1.9 |
|  | SI | 6.5 | 3.3 | 3.4 | 3.1 | 4.3 | 3.3 | 3.1 | 3.9 | 3.1 | 3.4 | 3.3 | 3.9 |
|  | SG | 10.1 | 6.0 | 5.9 | 5.9 | 7.0 | 7.0 | 6.0 | 6.9 | 5.9 | 5.9 | 6.4 | 6.7 |
| CB |  | 64 | 128 | 16 | 16 | 128 | 32 | 32 | 32 | 16 | 32 | 32 | 16 |
|  | RI | 2.7 | 1.4 | 0.6 | 0.6 | 1.2 | 0.8 | 0.5 | 0.6 | 0.4 | 0.5 | 0.5 | 0.7 |
| 7 | RG | 5.6 | 2.7 | 1.8 | 1.7 | 2.5 | 2.0 | 1.6 | 2.1 | 1.6 | 1.9 | 2.0 | 2.0 |
|  | SI | 6.9 | 6.0 | 3.4 | 3.7 | 4.3 | 3.8 | 3.8 | 3.8 | 3.3 | 3.4 | 3.5 | 3.9 |
|  | SG | 11.0 | 9.3 | 6.1 | 6.1 | 7.4 | 7.1 | 5.9 | 7.6 | 5.5 | 6.2 | 6.1 | 6.8 |
| CB |  | 32 | 64 | 8 | 32 | 64 | 32 | 16 | 16 | 16 | 16 | 32 | 16 |
|  | RI | 2.9 | 1.9 | 0.5 | 0.5 | 1.4 | 0.7 | 0.6 | 0.6 | 0.4 | 0.5 | 0.5 | 0.6 |
| 8 | RG | 5.3 | 3.4 | 1.7 | 1.9 | 2.4 | 3.1 | 2.6 | 3.1 | 1.6 | 1.8 | 2.0 | 2.0 |
|  | SI | 7.3 | 6.5 | 3.8 | 3.6 | 4.8 | 4.0 | 3.7 | 4.3 | 3.5 | 3.7 | 3.8 | 4.3 |
|  | SG | 11.1 | 9.7 | 6.3 | 6.6 | 8.0 | 8.6 | 7.0 | 8.4 | 6.3 | 6.5 | 6.8 | 7.3 |

The following system configurations were used with the test sets, from the results of this section:

- Single section VQ with codebook sizes of 64 and 128, since the best results were achieved.
- Multi-section VQ, for 2 (codebook sizes of 32 and 64) and 3 (codebook size of 32) sections, and Sum and Weighted Sum based on Random Forgery Error fusion techniques, since the best results were achieved with the multi-section.

___________________________________________________________________

It is important to emphasize that the computational burden is directly affected by codebook size, rather than the number of codebook sections. This is due to the direct proportionality between the number of vectors inside a codebook and the time required to find its nearest neighbor. For example, a multi-section approach of two sections and 4 bits per section is two times faster than a single codebook with 5 bits. In this case, the total number of vectors is the same ($2^4+2^4=2^5$), but the required time to find a nearest neighbor is only half the amount, because multi-section only requires $2^4$ vector distance computation for each vector to be quantized. In addition, a multi-section approach can model the time evolution, as it splits the signature into initial and final parts, while the single section mixes both parts.

4.1.2 Test sets

Tables 4 and 5 show the results with the TS1 (MCYT database) and TS2 (SVC corpus), respectively. The DS has been used to achieve the coefficient values for the WSRE fusion technique.

The results with the MCYT corpus are similar to those achieved with the development set, getting worse with random forgeries, which is common when the database increases in size. However, with the SVC database, the best results are achieved with the multi-section approach, more specifically with 2 sections, although the results with a single section are very still good.

From the results in the tables, it is difficult to choose an optimum system configuration. However, as can be seen in the next section, all the results achieved are very competitive with regard to those achieved with other proposals, while the system requirements are much lower.

**Table 4.** Results (EER in %) achieved with the test set of MCYT corpus. Best results are bold emphasized.

|  |  | **Random** |  | **Skilled** |  |
| --- | --- | --- | --- | --- | --- |
| **VQ Conf.** | **CB Size** | **Ind. Thres.** | **Gen. Thres.** | **Ind. Thres.** | **Gen. Thres.** |
| 1 Section | 64 | 0.26 % | **0.68 %** | 2.61 % | 5.56 % |
| 1 Section | 128 | **0.23 %** | 0.72 % | **2.46 %** | **4.92 %** |
| 2 S. (S) | 32 | 0.30 % | 0.83 % | 2.85 % | 5.76 % |
| 2 S. (WSRE) | 32 | 0.28 % | 0.73 % | 2.81 % | 6.06 % |
| 2 S. (S) | 64 | 0.28 % | 0.85 % | 2.65 % | 5.30 % |
| 2 S. (WSRE) | 64 | 0.27 % | 0.85 % | 2.66 % | 5.63 % |
| 3 S. (S) | 32 | 0.41 % | 0.92 % | 2.77 % | 5.57 % |
| 3S (WSRE) | 32 | 0.37 % | 0.86 % | 2.86 % | 5.58 % |

**Table 5.** Results (EER in %) achieved with the test set of SVC corpus. Best results are bold emphasized.

|  |  | **Random** |  | **Skilled** |  |
|---|---|---|---|---|---|
| VQ Conf. | CB Size | Ind. Thres. | Gen. Thres. | Ind. Thres. | Gen. Thres. |
| 1 Section | 64 | 0.038 % | **0.31 %** | 5.58 % | 17.29 % |
| 1 Section | 128 | 0.064 % | **0.31 %** | 5.50 % | 15.79 % |
| 2 S. (S) | 32 | **0.000 %** | **0.31 %** | 6.53 % | 16.85 % |
| 2 S. (WSRE) | 32 | **0.000 %** | 0.33 % | 6.52 % | 16.85 % |
| 2 S. (S) | 64 | **0.000 %** | 0.33 % | **5.00 %** | **15.50 %** |
| 2 S. (WSRE) | 64 | **0.000 %** | 0.33 % | 5.30 % | **15.50 %** |
| 3 S. (S) | 32 | 0.026 % | **0.31 %** | 6.17 % | 17.50 % |
| 3S (WSRE) | 32 | 0.026 % | **0.31 %** | 6.17 % | 17.50 % |

## 4.2 Identification task

### 4.2.1 Development set

Fig. 4 shows the identification rate with respect to the codebook size with a one section VQ. The same codebook sizes as in the verification task have been used.

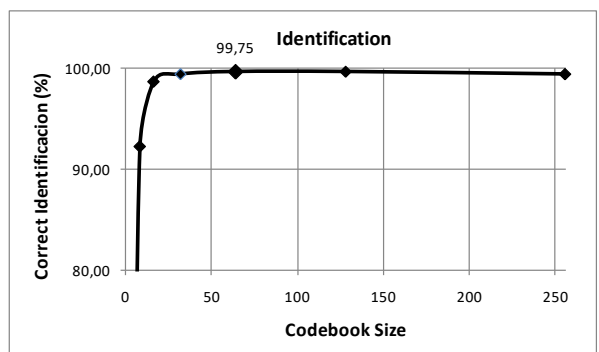

**Figure 4.** Identification results with the DS when a VQ with a section is used. The best result is shown.

Table 6 summarizes the results with the multi-section approach. The *Sec* column shows the number of sections. The other columns show the fusion techniques tested (see Sec. 2.1.3). The row CB shows the optimal codebook size (that with the best performance) for each number of sections and combination technique. The best results are emphasized in bold.

**Table 6.** Signatures correctly assigned (% of success) in the identification task obtained in the multi-section experiments with the development set.

|    | Sec | min   | max   | S     | P     | SEV   | WSD   | WSHM  | WSLM  | WSRE  | WSSE  | WSUE  | WSUT  |
|----|-----|-------|-------|-------|-------|-------|-------|-------|-------|-------|-------|-------|-------|
| CB |     | 32    | 16    | 32    | 32    | 32    | 64    | 32    | 32    | 64    | 32    | 64    | 32    |
|    | 2   | 99.25 | 98.50 | **99.75** | **99.75** | **99.75** | 99.5  | **99.75** | **99.75** | **99.75** | **99.75** | **99.75** | **99.75** |
| CB |     | 32    | 32    | 16    | 16    | 16    | 16    | 16    | 32    | 16    | 16    | 64    | 32    |
|    | 3   | 98.25 | 97.5  | **99.75** | 99.25 | 99.5  | 99.0  | 99.0  | 99.25 | **99.75** | **99.75** | 99.5  | 99.5  |
| CB |     | 64    | 16    | 16    | 32    | 16    | 16    | 16    | 64    | 32    | 32    | 32    | 32    |
|    | 4   | 94.75 | 97.75 | 99.0  | 99.0  | 98.25 | 97.75 | 98.75 | 98.5  | 99.5  | 99.0  | 98.5  | 98.5  |
| CB |     | 32    | 16    | 8     | 32    | 16    | 16    | 16    | 8     | 16    | 8     | 64    | 16    |
|    | 5   | 94.25 | 96.75 | 98.0  | 98.5  | 97.5  | 97.0  | 98.25 | 97.75 | 98.0  | 98.0  | 97.75 | 97.75 |
| CB |     | 32    | 32    | 8     | 8     | 32    | 16    | 16    | 16    | 16    | 8     | 64    | 16    |
|    | 6   | 94.75 | 96.25 | 98.0  | 98.25 | 97.25 | 98.0  | 98.5  | 98.0  | 98.0  | 97.75 | 98.25 | 98.25 |
| CB |     | 32    | 32    | 8     | 8     | 16    | 32    | 16    | 8     | 8     | 8     | 32    | 16    |
|    | 7   | 92.75 | 96.25 | 98.75 | 98.5  | 97.0  | 97.25 | 98.0  | 97.75 | 98.5  | 98.75 | 98.25 | 98.0  |
| CB |     | 16    | 8     | 8     | 8     | 16    | 8     | 8     | 4     | 4     | 4     | 64    | 16    |
|    | 8   | 90.0  | 94.25 | 98.0  | 98.25 | 95.5  | 95.75 | 96.75 | 96.25 | 98.0  | 98.0  | 98.0  | 97.75 |

From the results in Fig. 4 and table 6 we can conclude that, as in the verification task, the multi-section does not outperform the results with the single section. However, with multi-section, the same best results are achieved with fewer vectors in the codebook: 64 in the one section VQ, 32 in the two sections VQ and 16 in the three sections VQ; then, the smallest number of distances calculation (smallest CB) is achieved with the three sections VQ.

This tendency of getting good results with fewer vectors in the codebook, while the number of sections is increased, can also be seen in the rest of the results of table 6. It was therefore considered interesting to test the multi-section with the TS.

With regard to the fusion techniques, the *Sum*, *WSRE* and *WSSE* are the best. The speed in the response is very important in identification, so the simplest, the *Sum*, was used with the TS.

4.2.2 Test set

Fig. 5 shows the results with the Test Sets for codebook sizes 2, 4, 8, 16, 32 and 64 and 1, 2 and 3 sections for MCYT and 1, 2, 3 ,4 and 5 for SVC; greater sizes of both parameters do not outperform the results shown. The first thing to be emphasized is the very good system performance: **99.76%** of correct identification is achieved with the MCYT TS and **100%** with the SVC database.

The tendency seen with the DS is also observed here: the greater the number of sections, the smaller the codebook size needed to achieve better performance. Table 7 shows a summary of this, since the best results achieved with each number of sections and the corresponding codebook size are shown. As can be seen, the use of multi-section supposes a reduction in the number of distances calculation (proportional to the codebook size), thus improving the speed of the system without performance reduction with SVC, and with a very small one with MCYT.

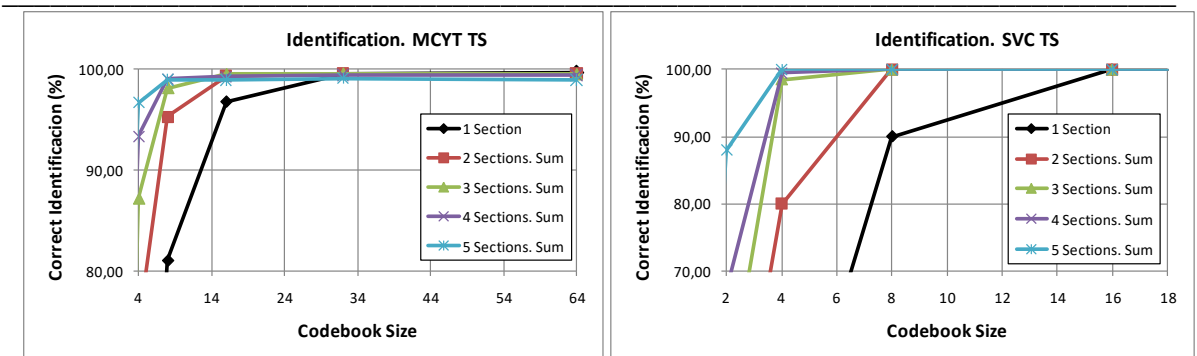

**Figure 5.** Identification results with the Test Sets.

**Table 7.** Best results achieved in identification with the TSs for section numbers 1 to 5. The corresponding codebook size also appears.

|  | 1 Section | | 2 Sections | | 3 Sections | | 4 Sections | | 5 Sections | |
|---|---|---|---|---|---|---|---|---|---|---|
|  | Id (%) | CB | Id (%) | CB | Id (%) | CB | Id (%) | CB | Id (%) | CB |
| MCYT | 99.76 | 128 | 99.68 | 32 | 99.60 | 32 | 99.28 | 16 | 99.12 | 32 |
| SVC | 100.0 | 16 | 100.0 | 8 | 100.0 | 8 | 100.0 | 8 | 100.0 | 4 |

### 4.3. State-of-the-art comparison

Tables 8 and 9 show some of the most recent published work with the MCYT and SVC databases, respectively. In these tables, the best performance achieved and brief descriptions of the main characteristics of each proposal are shown. Nevertheless, it is important to emphasize that straightforward comparison is not always possible. This is due to different training and testing conditions (which sometimes remain unclear) of different papers.

It can be seen that our system, in general, outperforms the state-of-the-art scores, more so taking into account the fact that the results shown in the tables are the best achieved in each work. Only very complex systems (Garcia-Salicetti 07), using fusion of classifiers and more features, outperform our proposals' performance. We think that our proposals' results would also improve if score normalization were applied. Although this is far from the scope of this work, it may be a interesting future work.

**Table 8** Performance achieved for other proposals with MCYT database. The reference to each work appears in the first column.

| Work | Classifier | Features | Exp. Environment | EER(%) |
|---|---|---|---|---|
| Fierrez 2005 | HMM (local expert) + Parzen Windows (global) | 7 + 7d | 330 users, 5 training signatures, I Thr | Ra: 0.2 Sk: 2.1 |
| Nanni 2005 | Ensemble of Parzen Windows | 60 best over 100 | 100 users, 5 training signatures, G Thr | Ra: 2.9 Sk: 8.4 |
| Nanni 2006 | Ensemble of one-class classifiers | 60 best over 100 | 100 users, 5 training signatures, G Thr | Ra: 2.1 Sk: 6.5 |
| Ketabdar 05 | GMM | 12 best over 150 | 50 users, 5 training signatures, G Thr | Sk: 4.5 |
| Hua 2006 | Based on FFT | 2 (*x* and *y*) | 100 users, 5 training signatures, G Thr | Sk: 7.0 |
| Faundez 07 | VQ+DTW | 5 | 280 users, 5 training signatures, G Thr | Ra: 1.4 Sk: 5.4 |
| Garcia 2007 | 4 expert fusion: 2 HMM+ 2 based on distances | 12-25 | 330 users, 5 training signatures, G Thr | Sk: 3.4 |
| Vivaracho09 | Normalized signatures + fractional distances | 5 | 330 users, 5 training signatures, I Thr | Ra: 0.5 Sk: 3.1 |

**Table 9** Performance achieved for other proposals with SVC database. The reference to each work appears in the first column.

| Work | Classifier | Features | Exp. Environment | EER(%) |
|---|---|---|---|---|
| Yeung 2004 | Best competition results | 2 | Task 1 of SVC, Individual Threshold | Ra: 3.49 Sk: 5.50 |
| Yeung 2004 | Best competition results | 5 | Task 2 of SVC, Individual Threshold | Ra: 3.02 Sk: 6.90 |
| Fierrez 2005 | DTW + HMM | 3 (local), 7 (regional) | Task 2 of SVC, Individual Threshold | Ra: 0.15 Sk: 6.91 |
| Van Bao 07 | HMM likelihood information | 25 | Task 2 of SVC, Individual Threshold | Sk: 4.83 |
| Gruber 2009 | SVM, LCSS Kernel | 5 | Task 2 of SVC, Individual Threshold | Ra: 0.12 Sk: 6.84 |

## 5. CONCLUSIONS

In this paper, we have proposed several VQ based approaches for on-line signature recognition. These algorithms enable us to take into account the temporal evolution of the signature, and obtain a significant improvement in speed when compared with the state-of-the-art algorithms. This improvement is due to the lower computational burden of the VQ approach, which has been neglected for signature recognition so far; although it has proven useful in the past for other biometric traits, such as speech, especially for short training and testing sets.

Experimental results on MCYT and test sets reveal a very competitive performance:

- Identification rates up to 99.76%.
- EER equal to 0.23% (individual threshold) and 0.68% (general threshold) for random forgeries.
- EER equal to 2.46% (individual threshold) and 4.92% (general threshold) for skilled forgeries.

___________________________________________________________________

Experimental results on SVC reveal the same competitive results:

- Identification rates up to 100%.
- EER equal to 0.000% (individual threshold) and 0.31% (general threshold) for random forgeries.
- EER equal to 5% (individual threshold) and 15.5% (general threshold) for skilled forgeries.

In addition, our system improves the database storage requirements due to vector compression, and is more privacy-friendly, as it is not possible to recover the original signature using the codebooks.

In the appendixes, we demonstrate that VQ is around 47 times faster than the classic DTW algorithm, which provided verification errors of 1.4% (random forgeries) and 5.4% (skilled forgeries) over the MCYT database, when using a general threshold (Faundez-Zanuy, 07).

## APPENDIX 1: VQ Algorithm

Given a distance measure between vectors *i* and *j*, such as, for instance, the Euclidean distance:
$$D(i,j) = \|i-j\|_2 \tag{4}$$
The distance between a codebook and a candidate's signature can be computed using the following algorithm:

*Initialization:*
$\tilde{D} = 0$

*Recursion:*
For $i=1, \ldots, I$
$\quad d_{min} = D(i,1)$
For $j=2, \ldots, L$
$\quad d = D(i,j)$
$\quad if \quad d < d_{min} \Rightarrow d_{min} = d$
End
$\tilde{D} = \tilde{D} + d_{min}$
End

*Termination*

The best match has a cost of $\tilde{D}$.

## APPENDIX 2: DTW Algorithm

To find the optimal path to $(i_k, j_k)$, we simply take the minimum over-all distance predecessors:
$$D_{min}(i_k, j_k) = \min_{(i_{k-p}, j_{k-p})} \{D_{min}[(i_k, j_k)|(i_{k-p}, j_{k-p})]\} \tag{1}$$
The simplest case corresponds to three predecessors.

___

The distance between a reference signature and a candidate's signature can be computed using the following algorithm:

*Initialization:*

$$D_{\min}(1,j) = d_N(1,j), \quad j=1,\cdots,\varepsilon$$
$$D_{\min}(i,1) = d_N(i,1), \quad i=2,\cdots,\varepsilon$$
$$\delta_1(j) = D_{\min}(1,j), \quad j=1,\cdots,J$$
$$\delta_2(j) = 0, \quad j=1,\cdots,J$$

*Recursion:*

    For $i=2,\ldots,I$
        For $j=J,\ldots,2$
            Compute $D_{\min}(i,j)$ using equation (1)
$$\delta_2(j) = \delta_1(j)$$
$$\delta_1(j) = D_{\min}(i,j)$$
        Next *j*
    Next *i*

*Termination*

The best path has cost:
$$\tilde{D} = \min \begin{cases} D_{\min}(I,j)/I, & j = J-\varepsilon,\ldots,J \\ D_{\min}(i,J)/I, & i = I-\varepsilon,\ldots,I \end{cases}$$

This notation is consistent with that provided in (Deller et al., 1987). A complete explanation of this dynamic programming technique is beyond the scope of this paper.

**APPENDIX 3: COMPUTATIONAL BURDEN COMPARISON**

In this appendix we compare the computational burden of the DTW algorithm and the proposed codebook approach. We use the following nomenclature:

- *J* is the average single-signature reference template length.
- *I* is the candidate's signature length.
- *K* is the number of reference templates per user.
- *L* is the number of vectors inside the codebook for the VQ approach.
- *S* is the number of sections in the multi-section VQ approach.

In our experiments, we have set *K*=5, and in our database (MCYT), the average length per signature is *J*=454 vectors.

It is interesting to observe that, due to the vector quantization of the *K* reference templates per user, the number of reference vectors has been significantly reduced for the VQ algorithm, since all the reference signatures have been clustered together in a single codebook per user. For a codebook of 4 to 7 bits, we get *L*=16, 32, 64 and 128 vectors respectively, while the original average number of vectors per signature is 454. In addition, for DTW, the procedure must be executed for each reference signature per user (in our experimental data we have used

___

$K$=5). Thus, even for a 7 bit codebook, the VQ approach requires approximately 18 times (5×454/128) less data to be dealt with.

Dynamic time warping requires the computation of $KIJ$ distance measures. However, the search region can be restricted to a parallelogram region with slopes ½ and 2. Search over this parallelogram requires about $O(KIJ/3)$ distance measures to be computed and the DP equation (1) (see appendix 1) to be used about $O(KIJ/3)$ times. This latter figure is often referred to as the "number of DP searches" (Deller et al., 1987).

VQ requires the computation of O($IL$) distance measures. It is interesting to observe that the number of computations is the same for VQ and multi-section VQ, because the unique difference between them is the change of codebook, depending upon which section a given vector belongs to. Taking into account that each DTW distance computation requires the computation of at least three distances between vectors, we can establish that VQ is approximately 47 times faster than DTW (for a codebook of 4 bits).

In terms of database storage requirements, DTW implies the storage of the whole set of reference signatures, which implies $KJ$ vectors per user. VQ requires $L$ vectors per user, where $L$ is the number of vectors inside the codebook, and this figure must be increased by the number of sections for the multi-section VQ approach.

The table 10 summarizes the computational and database memory requirements.

**Table 10.** Database storage and computational requirements for DTW and VQ approaches. It is important to emphasize that optimal L values are smaller for Multi-section VQ than for VQ (single section). Typically,

$$L_{VQmulti-section} = \frac{L_{VQ}}{S}$$

| Requirements | DTW | VQ | Multi-section VQ |
|---|---|---|---|
| Database memory | $KJ$ | $L$ | $SL$ |
| Computational (Number of distance computations) | $O(KIJ/3)$ | $O(IL)$ | $O(IL)$ |

**REFERENCES**

Buck, J. T., Burton, D. K., Shore, J. E., 1985. Text-dependent speaker recognition using vector quantization. IEEE ICASSP 1985, pp.391-394

Burton, D. K., Shore, J. E., Buck, J. T., 1983. A generalization of isolated word recognition using vector quantization. IEEE ICASSP 1983 Boston, pp. 1021- 1024

Deller, J. R., Proakis, J. G., Hansen, J. H. L., 1987. Discrete-time processing of speech signals. Ed. Prentice Hall.

Faundez-Zanuy, M., 2004. On the vulnerability of biometric security systems. IEEE Aerospace and Electronic Systems Magazine, vol. 19, no. 6, pp. 3-8, June 2004, doi: 10.1109/MAES.2004.1308819

Faundez-Zanuy M., 2005. Data fusion in biometrics. IEEE Aerospace and Electronic Systems Magazine, vol. 20, no. 1, pp. 34-38, Jan. 2005, doi: 10.1109/MAES.2005.1396793

Juan Manuel Pascual-Gaspar, Marcos Faundez-Zanuy, Carlos Vivaracho "Fast on-line signature recognition based on VQ with time modeling", Engineering Applications of Artificial Intelligence, Volume 24, Issue 2, 2011, Pages 368-377, ISSN 0952-1976, https://doi.org/10.1016/j.engappai.2010.10.015 .

___________________________________________________________________

Faundez-Zanuy M., 2005b. Privacy issues on biometric systems. IEEE Aerospace and Electronic Systems Magazine. IEEE Aerospace and Electronic Systems Magazine 20(2):13 – 15, DOI: 10.1109/MAES.2005.1397143

Faundez-Zanuy, M., 2005c. Biometric recognition: why not massively adopted yet?. IEEE Aerospace and Electronic Systems Magazine, vol. 20, no. 8, pp. 25-28, Aug. 2005, doi: 10.1109/MAES.2005.1499300.

Faundez-Zanuy, M., Monte-Moreno, E., 2005d. State-of-the-art in speaker recognition. IEEE Aerospace and Electronic Systems Magazine, vol. 20, no. 5, pp. 7-12, May 2005, doi: 10.1109/MAES.2005.1432568.

Faundez-Zanuy, M., 2007. On-line signature recognition based on VQ-DTW. Pattern Recognition, Volume 40, Issue 3, March 2007, Pages 981-992, https://doi.org/10.1016/j.patcog.2006.06.007

Fierrez-Aguilar, J. Nanni, L., Lopez-Peñalba, J., Ortega-Garcia, J., Maltoni, D., 2005. An on-line signature verification system based on fusion of local and global information., in: Audio- and Video-Based Biometric Person Authentication, vol. 3546/2005 of Lecture Notes in Computer Science, Springer Berlin / Heidelberg, pp. 523-532.

Frias-Martinez, E., Sanchez, A. and Velez, J., 2006. Support vector machines versus multi-layer perceptrons for efficient off-line signature recognition. Engineering Applications of Artificial Intelligence, 19 (6), pp. 693-704, Special Section on Innovative Production Machines and Systems (I*PROMS), September.

Garcia-Salicetti, S., Fierrez-Aguilar, J., Alonso-Fernandez, F., Vielhauer, C., Guest, R., Allano, L., Trung, T. D., Scheidat, T., Van, B. L., Dittmann, J., Dorizzi, B., Ortega-Garcia, J. Gonzalez-Rodriguez, J., di Castiglione, M. B., Fairhurst, M., 2007. Biosecure reference systems for on-line signature verification: A study of complementarity. Annals of Telecommunications, Special Issue on Multimodal Biometrics 62 (1-2), pp. 36-6.

Gersho, A., Gray, R.M., 1991. Vector Quantization and Signal Compression. Kluwer Academic Publishers.

Guyon, I., Makhoul, J., Schwartz, R., Vapnik, V., 1998. What size test set gives good error rate estimates?'. IEEE Trans. Pattern Anal. Mach. Intell., 1998, 20, (1), pp. 52–64

Han, K., Sethi, I. K., 1996. Handwritten signature retrieval and identification. Pattern recognition letters, vo. 17, pp. 83-90.

Higgins, A. L., Bahler, L. G., Porter, J. E., 1993. Voice identification using nearest –neighbor distance measure". IEEE International Conference on Acoustics, Speech and Signal Processing ICASSP 1993. pp. II-375, II-378

N. Houmani, A. Mayoue, S. Garcia-Salicetti, B. Dorizzi, M.I. Khalil, M.N. Moustafa, H. Abbas, D. Muramatsu, B. Yanikoglu, A. Kholmatov, M. Martinez-Diaz, J. Fierrez, J. Ortega-Garcia, J. Roure Alcobé, J. Fabregas, M. Faundez-Zanuy, J.M. Pascual-Gaspar, V. Cardeñoso-Payo, C. Vivaracho-Pascual, "BioSecure Signature Evaluation Campaign 2009 (BSEC'2009): Results" Pattern Recognition, Volume 45, Issue 3, 2012, Pages 993-1003, https://doi.org/10.1016/j.patcog.2011.08.008

Hua Quan, Z., Shuang Huang, D., lei Xia, X., Lyu, M., Lok, T., 2006. Spectrum analysis based on windows with variable widths for online signature verification. In: ICPR 2006, 18th International Conference on, vol. 2, pp. 1122-1125.

Juan Manuel Pascual-Gaspar, Marcos Faundez-Zanuy, Carlos Vivaracho "Fast on-line signature recognition based on VQ with time modeling", Engineering Applications of Artificial Intelligence, Volume 24, Issue 2, 2011, Pages 368-377, ISSN 0952-1976, https://doi.org/10.1016/j.engappai.2010.10.015 .

___________________________________________________________________

Ketabdar, H., Richiardi, J., Drygajlo, A., 2005. Global feature selection for on-line signature verification. In: Proc. 12th Internat. Graphonomics Society Conf.

Ortega-Garcia, J., Gonzalez-Rodriguez, J., Simon-Zorita, D., Cruz-Llanas S., 2002. From biometrics technology to applications regarding face, voice, signature and fingerprint recognition systems. Chapter 12, pp. 289-337 in Biometrics Solutions for Authentication in an E-World. Kluwer Academic Publishers.

Jain, A. , Bolle, R., Pankanti, S., 1999. Biometrics. Personal identification in a networked society. Kluwer Academic Publishers.

Jain, A. K., Duin, R. P. W., Mao, J., 2000. Statistical pattern recognition: a review. IEEE Transactions on Pattern Analysis and Machine Intelligence. Vol. 22 No 1, January.

Martin A. et al., 1997. "The DET curve in assessment of detection performance", V. 4, pp.1895-1898, European speech Processing Conference Eurospeech.

Nanavati, S., Thieme, M., Nanavati, R., Biometrics., 2002. Identity verification in a networked world. John Wiley & sons.

Nanni, L., Lumini, A., 2005. Ensemble of parzen window classifiers for on-line signature verification. Neurocomputing 68, pp., 217-224.

Nanni, L., 2006. Experimental comparison of one-class classifiers for online signature verification. Neurocomputing 69 (7-9), pp. 869-873.

Ortega-Garcia, J., Fierrez, J., Simon, D., Gonzalez, J., Faundez-Zanuy, M., Espinosa, V., Satue, A., Hernaez, I., Igarza, J.-J. , Vivaracho, C., Escudero, D., Moro, Q.-I., 2003. MCYT Baseline Corpus: A Multimodal Biometric Database. IEE Proceedings - Vision, Image and Signal Processing Vol. 150, pp.395-401, December. DOI:  10.1049/ip-vis:20031078

Pascual-Gaspar, J. M., Cardeñoso-Payo, V., Vivaracho-Pascual, C., 2009. Practical On-Line Signature Verification, LNCS/LNAI, Advances in Biometrics, 3th International Conference, ICB 2009, n. 5558, pp. 1180-1189.

Plamondon, R. , Lorette, G., 1989. Automatic signature verification and writer identification-the state of the art, Pattern Recognition, vol. 1, no. 2, pp.107-131.

Soong, F., Rosenberg, A., Rabiner, L., Juang, B., 1985. A vector Quantization approach to Speaker Recognition. IEEE Proceedings International Conference on Acoustics, Speech and Signal Processing ICASSP 1985 Vol.1 pp.387-390.Tampa

Taylan Das, M. and Canan Dulger, L., 2009. Signature verification (SV) toolbox: Application of PSO-NN. Engineering Applications of Artificial Intelligence, 22 (4-5), pp. 688-694, June.

Van Bao,  L., Garcia-Salicetti, S. and Dorizzi, B., 2007. On using the Viterbi path along with HMM likelihood information for online signature verification. IEEE Transactions on Systems,Man, and Cybernetics, Part B, 37, 1237–1247.

Vivaracho, Carlos E., Ortega-Garcia, Javier, Alonso, Luis, y Moro, Quiliano I., 2003. Improving  the competitiveness of discriminant neural networks in speaker verification. Proc. Eurospeech, ISSN 1018-4074, pp. 2637-2640, September.

Vivaracho-Pascual, C., Faundez-Zanuy, M. and Pascual, J. M., 2009. An efficient low cost approach for on-line signature recognition based on length normalization and fractional

___

distances. Pattern Recognition, 42(1), pp. 183-193. https://doi.org/10.1016/j.patcog.2008.07.008

Xiao, X., Leedham, G., 2002. Signature verification using a modified Bayesian network. Pattern Recognition Vol. 35, Issue 5, May, pp.983-995.

Yeung, D., Chang, H., Xiong, Y., George, S., Kashi, R., Matsumoto, T., Rigoll, G., 2004. SVC2004: First international signature verification competition. Lecture Notes on Computer Science LNCS-3072, Springer Verlag pp.16-22.

Zhang, D., 2000. Automated biometrics. Technologies and systems. Kluwer Academic Publlishers.